\documentclass[sigconf]{acmart}
\usepackage{fancyvrb}

\usepackage{booktabs}

\def\BibTeX{{\rm B\kern-.05em{\sc i\kern-.025em b}\kern-.08emT\kern-.1667em\lower.7ex\hbox{E}\kern-.125emX}}
  
\copyrightyear{2019}
\acmYear{2019}
\setcopyright{acmlicensed}
\acmConference[FEED 2019]{Fragile Earth: Theory Guided Data Science to Enhance Scientific Discovery, KDD Workshop}{August 5, 2019}{Anchorage, Alaska}
\acmConference[]{}
\acmPrice{}
\acmDOI{}
\acmISBN{}
\renewcommand\footnotetextcopyrightpermission[1]{}  
\begin{document}
\title{Accelerating Physics-Based Simulations Using Neural Network Proxies: An Application in Oil Reservoir Modeling }
\author{Ji\v{r}\'i Navr\'atil}
\email{jiri@us.ibm.com}
\affiliation{%
  \institution{IBM Research}
  \city{Yorktown Heights}
  \state{New York}
  \postcode{10598}
}
\author{Alan J. King}
\email{kingaj@us.ibm.com}
\affiliation{%
  \institution{IBM Research}
  \city{Yorktown Heights}
  \state{New York}
  \postcode{10598}
}

\author{Jesus Rios}
\email{jriosal@us.ibm.com}
\affiliation{%
  \institution{IBM Research}
  \city{Yorktown Heights}
  \state{New York}
  \postcode{10598}
}

\author{Georgios Kollias}
\email{gkollias@us.ibm.com}
\affiliation{%
  \institution{IBM Research}
  \city{Yorktown Heights}
  \state{New York}
  \postcode{10598}
}

\author{Ruben Torrado}
\authornote{Author is currently with OriGen.AI, New York, NY  }
\email{rubentorrado@gmail.com}
\affiliation{%
  \institution{Repsol, S.A.}
  \streetaddress{Mostoles}
  \city{Mostoles}
  \country{Spain}
}
\author{Andr\'es Codas}
\email{andrescodas@br.ibm.com}
\affiliation{%
  \institution{IBM Research}
  \city{Rio De Janeiro}
  \country{Brazil}
}

\renewcommand{\shortauthors}{Navr\'atil et al.}
\renewcommand{\shorttitle}{Accelerating Physics-Based Simulations Using Neural Network Proxies}

\begin{abstract}
We develop a proxy model based on deep learning methods to accelerate the simulations of oil reservoirs--by three orders of magnitude--compared to industry-strength physics-based PDE solvers. This paper describes a new architectural approach to this task, accompanied by a thorough experimental evaluation on a publicly available reservoir model. We demonstrate that in a practical setting a speedup of more than 2000X can be achieved with an average sequence error of about 10\% relative to the oil-field simulator. 
The proxy model is contrasted with a high-quality physics-based acceleration baseline and is shown to outperform it by several orders of magnitude. We believe the outcomes presented here are extremely promising and offer a valuable benchmark for continuing research in oil field development optimization.
Due to its domain-agnostic architecture, the presented approach can be extended to many applications beyond the field of oil and gas exploration. 
\end{abstract}
\maketitle
\section{Introduction} \label{Sec:Introduction}
Reservoir modeling plays an essential role in modern oil and gas exploration. After an acquisition of a reservoir field, energy companies plan the field development based on a capital expense of billions of dollars. A placement of even a single well in a bad spot can mean a significant economical loss for the developer. 
A reservoir model (RM) is a computerized representation of the field drawing from various data sources, such as geological expert analyses, seismic measurements, well logs, etc., with added properties determining the dynamic reservoir behavior.
Its primary purpose is to allow optimization and better prediction of the field's future output using mathematical simulators. Given a set of input actions (well drilling), a simulator operates on an RM by solving large systems of non-linear PDEs to predict future outcomes over long time horizons (up to tens of years). 

RM simulators may require a considerable amount of computation (time) to produce predictions (minutes, hours, or days, depending on the RM size). Current optimization techniques in reservoir engineering are therefore able to simulate only a small number of cases. 
In our context, the problem of field development is formulated as an optimization over a very high number (millions) of candidate sequential well placements. As such, the solving time of a simulation quickly becomes the bottleneck rendering methods such as Monte Carlo Planning and  Reinforcement Learning impractical. While the idea of creating an approximating surrogate, or proxy, is not new, previous approaches targeted accelerating steps inside the physics-based simulator (see Section \ref{Sec:PreviousWork}). Instead, we pursue a deep learning approach to learn approximating essential output variables of a PDE-solver using training data generated by that solver, and demonstrate its performance and accuracy at predicting production rates on a benchmark RM with complex geological properties, namely the SPE9 \cite{JR:SPE9}.

\section{Previous Work} \label{Sec:PreviousWork}
Accelerating reservoir simulations has a wide literature.  
State of the art techniques can be classified in two categories: 1) reducing the complexity of the PDEs while providing an acceptable loss of prediction accuracy (e.g., 
reduced order modeling \cite{IJNME:Durlofsky} can accelerate the simulations by factors of $10^2$ and has been found useful in optimization and control \cite{OptE:Durlofsky}, and upscaling which achieves acceleration with coarser reservoir models \cite{Durlofsky2005}),
and (2) simple polynomial interpolation techniques computing the objective function (e.g. Net Present Value, or NPV) or to characterize the uncertainty \cite{valladao2013stochastic}. 
It is important to note that while the latter techniques are fast their accuracy tends to be poor. 

Artificial Neural Networks (ANN) have been proposed to accelerate oil reservoir simulations and aim to achieve both objectives: accuracy and acceleration.
A special issue of \emph{Journal of Petroleum Science and Engineering} in 2014 was devoted to this topic, and a good summary can be found in the editorial \cite{JPSE:editorial}.  As outlined in this summary, ANN approaches can be categorized as (1) physics-based models, and  (2) data-driven models. 
Physics-based models use the PDE structures as features. Recent example from this category is application of Convolutional Neural Networks (CNN) to modeling flow around complex boundaries for the purpose of accelerating animation \cite{Google}. 
In our preliminary investigation, CNNs gave acceptable results modeling short time horizons, however, significant error accumulation over longer time spans of a reservoir model made this approach impractical.
The challenge to purely data-driven approaches is the extreme nonlinearity of the reservoir dynamics.  It seems important to incorporate features of the reservoir.  Examples of basic, fully connected ANNs to predict reservoir production are \cite{SPE:Sandstone} and \cite{IFP:Gravity}. In the context of pre-existing literature, we believe our contribution is twofold: (1) our architectural approach is unique as it deals with sequential action input and output of varying span, reservoir uncertainty, and optimized well control, and (2) we present a thorough experimental analysis on a publicly available reservoir model thus creating a valuable reference for future comparison by the community.

\section{Reservoir Model Simulation} \label{Sec:ReservoirModelAndSimulation} 
The Reservoir Model (RM) simulation considered in this work is based on a so-called Black Oil model \cite{trangenstein_bell_1989} that describes the flow of reservoir three fluids through porous rock: oil, water, and gas.
\subsection{Physics} \label{Sec:Physics}
The Black Oil model equates the time change of mass in a region with the mass flux across the region boundary. The flux is driven by pressure differences caused by well operations. 
The system of equations is highly nonlinear due to the non-stationary interactions between rock types and fluids.  Rock compressibility and relative permeability are altered by changes in fluid pressure and saturation levels. 
In addition, petroleum fluids will undergo phase changes, between gas and liquid form, as they move through pressure and temperature gradients.
An accessible introduction into the model and inner-workings of Black Oil Simulators (BOS) can be found in \cite{MRSTBook}. 

A BOS is a basic discrete-time finite-volume simulation.  The reservoir is partitioned into cells and the values of the primary variables: oil pressure, water saturation, and gas saturation, which are evaluated over a sequence of time steps out to a fixed time horizon. 

At each time step, the BOS solves the mass balance equations:
\begin{equation}
    \frac{ M_{c,f} \left( x_t \right ) - M_{c,f} \left ( x_{t-1} \right ) } { \Delta t }
    	= F_{c,f} \left ( x_t \right )
    	+ Q_{c,f} \left ( x_t, u_t \right ) 
    \label{Eq:SimulatorEquations}
\end{equation}
for each cell $c$ and each fluid $f$, where
    \begin{equation}
    \begin{array}{ll}
    x_t                      	& \textnormal{-- Field properties $[p_o,s_w,s_g]$}\\
    M \left ( x_t \right ) 		& \textnormal{-- Fluid mass in cell} \\
    F \left ( x_t \right ) 		& \textnormal{-- Mass flow/to from neighbor cells} \\
    u_t							& \textnormal{-- Well controls} \\
    Q \left ( x_t, u_t \right ) 	& \textnormal{-- Well flows }
    \end{array}
    \end{equation}
In a real reservoir model there may be up to $10^6$ cells and $10^3$ time steps, so the number of \emph{nonlinear} equations to be solved simultaneously can be on the order of $10^9$, although some practical problems are smaller.
Depending on the degree of the nonlinearity and the length of the time step, 
the set of equations to be solved for each time step can 
require many iterations of a nonlinear equation solver.  
A high-fidelity reservoir simulation for industrial applications can take hours or days to complete one simulation.

\subsection{Reservoir Uncertainty} \label{Sec:Realizations}
Each cell has a rock type, for example sandstone or shale, which defines important properties, such as compressibility and permeability.  However, due to the depth of reservoirs and the complex geology surrounding their formation, the attribution of a rock type to any given cell is somewhat uncertain.

An important notion is one of a {\em realization}, representing a particular 
spatial distribution of rock type across the grid cells of the RM - a distribution that comes with a natural uncertainty \cite{MRSTBook}.
This can be modeled by three-dimensional probability density functions, called variograms, that are calibrated from geological knowledge, seismic data, and cores from test wells.  In a typical application,  analysts will work with a few hundred samples (referred to as {\em Realizations}). 
To take this uncertainty into account in our experimental setup, we have used 500 realizations generated from the original RM according to \cite{caers2004multiple} with subsequent porosity and permeability calculation described in \cite{mariethoz2014multiple}.

\subsection{Wells} \label{Sec:Wells}
Another important element in BOS are {\em wells}. They have two functions: (1) the production of commercially valuable fluids (referred to as {\em Producer} wells), and (2)  the management of pressure differentials and fluid saturations in the reservoir.  
Some wells are drilled specifically to inject water or gas (referred to as {\em Injectors}).

The simulator predicts the impact of well operations which consist of two types of decisions: (1) the location and time sequence of well completions, and (2) the control of the production or injection rate.  The ultimate goal is to maximize the expected NPV.
\subsection{BOS Implementations} \label{Sec:BOSImplementations}

Our work relies on the following BOS packages: (1) \emph{Open Porous Media} (OPM) - an open source project supported by a consortium of companies and academic institutions \cite{JR:OPM}, and (2) \emph{Eclipse} \cite{EclipseURL} - a commercial software.

\section{SPE Benchmarks} \label{Sec:SPEBenchmarks}
\begin{figure}[htbp!]
       \centering
       \includegraphics[width=0.4\textwidth, height=3.5cm]{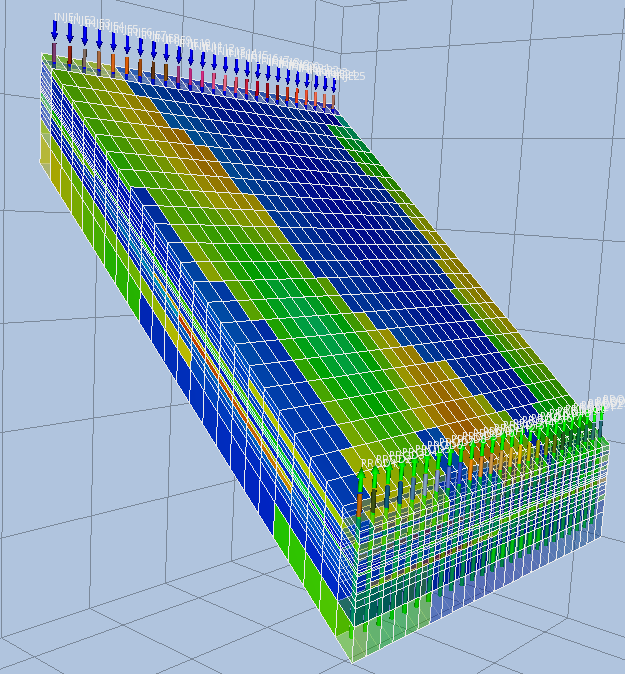}
       \caption{An example realization of the SPE9 RM (water pressure distribution).}
       \label{Fig:SPE9Reservoir}
\end{figure}

Our study draws data from one (of several) RM considered reference within the oil and gas industry: the SPE9 model \cite{JR:killough:1995,JR:SPE9}. 
The model consists of a $24 \times 25 \times  15$ grid and we used a varying number of injector and producer wells as will be described later. One particular realization of the SPE9 RM is shown in Figure \ref{Fig:SPE9Reservoir} illustrating the shape (and incline), grid structure and property (water pressure in this example) distribution within the RM.

\section{Learning to Predict Production Rates - The Proxy Approach} \label{Sec:LearningToPredictProductionRates}

\subsection{Task} \label{Sec:Task}
The main targets for the proxy model to predict are a set of variables referred to as {\em production rates}, taking into account interactions between wells, fluid, and rock. More specifically, given a particular RM realization and a series of {\em drilling actions}, a simulator predicts rates at which the wells produce or inject fluids at future times (referred to as time steps). A major application of such simulations is in Field Development Planning \cite{FDP_jahn_cook_graham_2011}, where the total production rates constitute economic metrics, such as the Net Present Value (NPV). Individual rates of each well drilled may also be of importance and we will include these in our investigations as well. 

Considering the above, we define the task of proxy modeling as follows:
\begin{itemize}
\item[] {\em Given}: 
  \begin{enumerate}
  \item Sequence of actions. An action encapsulates a drilling decision, in our case one of \{"Drill a Producer Well (P)", "Drill an Injector Well (I)", "Do Nothing (X)"\}, accompanied by applicable coordinates on the reservoir surface. 
  \item Realization ID. This ID maps to a known distribution of rock properties in the reservoir.
  \end{enumerate}
\item[] {\em Predict}: 
  \begin{enumerate}
  \item Field rates, i.e., aggregate output of the entire field, typically including oil, and water production as well as water injection over a desired future horizon in (equidistant) time steps.
  \item (Optional) Individual well rates, i.e., individual output of each producer well placed so far, typically oil production is of interest
  \end{enumerate}
\end{itemize}
An example of an action sequence suitably encoded could be as follows:
\begin{center}
\begin{BVerbatim}
x-P(0,15)-I(2,5)-x-x-P(23,19)-...-x
\end{BVerbatim}
\end{center}
where $\mbox{\texttt{P(0,15)}}$ refers to drilling a Producer at surface location (0,15), etc.
Continuing on this example, an expected output could be a series of rates predicted at increasing time steps, each representing a 30-day interval. In our experiments we used action sequences of length 20, and varying prediction spans (horizon) of 20 to 40 months. 

\subsection{Model}
\subsubsection{Encoder-Decoder Architecture }
\begin{figure}[htbp!]
       \centering
       \includegraphics[height=5.7cm]{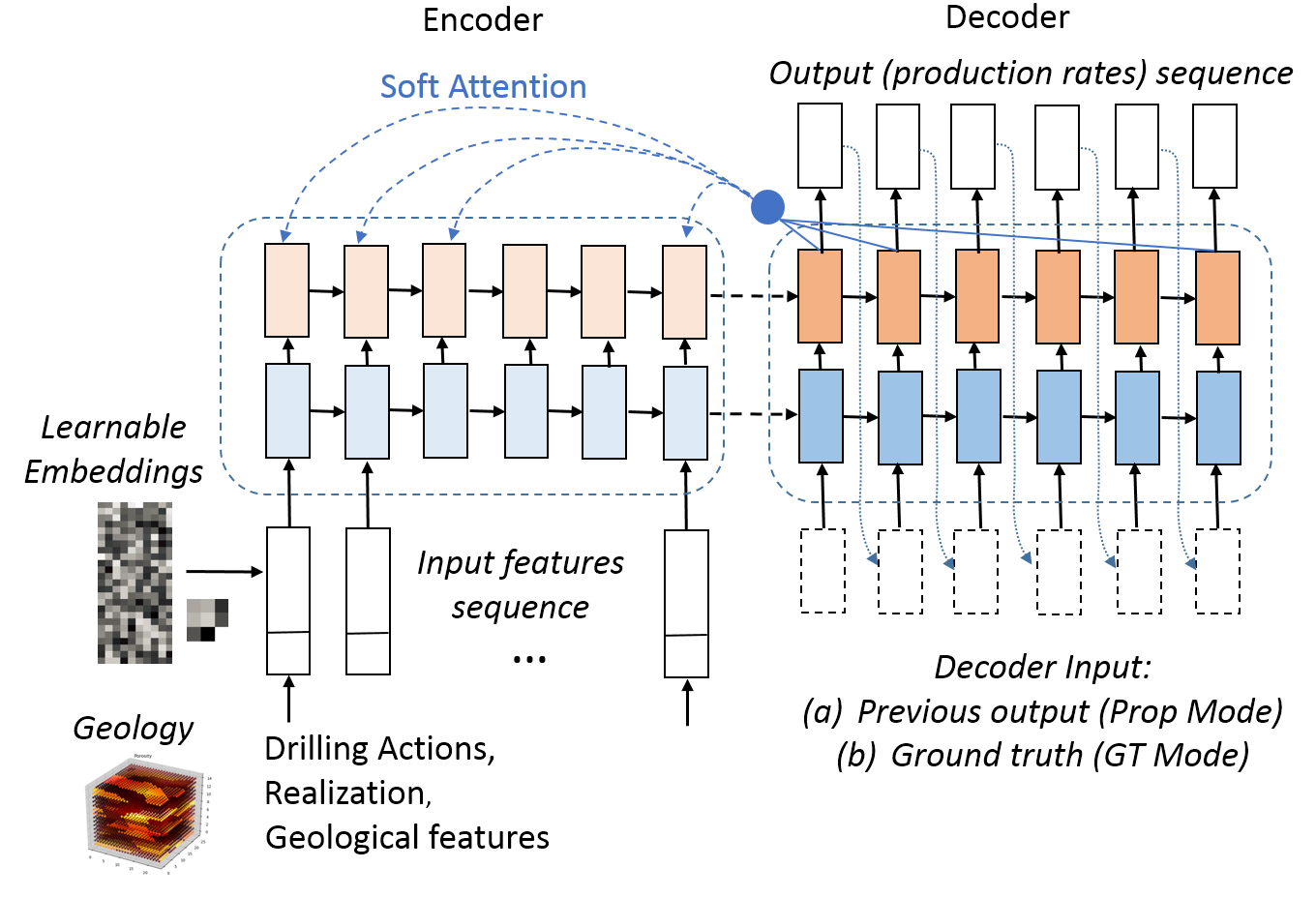}
       \caption{An architecture overview of the model}
       \label{Fig:EncDecArchitecture}
\end{figure}

Given the above task definition, our method adopts the sequence-to-sequence approach based on recurrent neural networks (RNNs), specifically a version of a gated RNN known as Long-Short Term Memory (LTSM) cell \cite{HochreiterLSTM}, arranged in a encoder-decoder architecture. This approach has gained popularity across various application fields \cite{SutskeverEncDecNLP,GravesEncDecSpeech,VinyalsEncDecImageCaptioning} and excels at modeling temporal sequences of (typically discrete) variables. In our setting, however, the model performs a continuous-variable regression, as will be explained more in detail below. Figure \ref{Fig:EncDecArchitecture} illustrates our solution. In this view, the series of drilling actions, along with additional geological information is the input sequence, $X = \{x_i\}_{0\leq i<K}$, while the series of production rates is the output sequence, $Y = \{y_t\}_{0\leq t<T}, \, y_t\in\mathbb{R}^D$, where $D$ is the number of fluid phases. Here, $x_t$ represents a union of potentially heterogeneous features, such as discrete and multivariate continuous variables. Later we will discuss how $X$ is transformed into a common space via embeddings.

While typical encoder-decoder applications deal with a discrete output defined over a closed set (of classes), i.e., modeling class probabilities with a training objective being a likelihood function (cross-entropy), 
in our setting the model performs a regression in $\mathbb{R}^D$ and is trained with the objective of minimizing the mean squared error (MSE).

The role of the Encoder is to capture the information about the input $X$ and pass it to the Decoder to generate an accurate prediction $\hat{Y}$ about $Y$. Both the encoder and the decoder involve multiple layers of LSTMs in our architecture.

\subsubsection{Neural Attention Mechanism}
Our model adopts the attention mechanism introduced in \cite{BahdanauCB14} to aid modeling the causal nature between the actions and the output. In order to generate a prediction $\hat{y}_t$, the attention mechanism applies a probability distribution (a "mask") over the individual input actions to emphasize actions particularly relevant to produce $y_t$. 
The mask itself is generated by a layer trained jointly with the rest of the network parameters. 
In contrast to a typical setup where the attention is applied on the upper-most encoder layer, we have found that concatenating encoder states from all layers outperformed the former. 

\subsubsection{Decoding Variants} \label{Sec:DecodingVariants}
We investigate the following modes of operation with respect to the decoder:
\begin{itemize}
\item Ground Truth (GT) Training, in which the decoder is provided the GT, i.e., $\{y_t\}_{0\leq t<T}$ as input at each time step. 
\item Prediction Propagation Training (or {\em Prop Training}), where in order to generate $\hat{y}_t$, the decoder uses its own prediction $\hat{y}_{t-1}$ as input (dotted decoder connections in Figure \ref{Fig:EncDecArchitecture})
\item {\em GT Pre-Training}, in which the model is first trained in GT Training mode, followed by Prop Training. This relates to an idea proposed in \cite{BengioNIPS15}.
\item Hybrid Propagation (or {\em HybridProp}), in which first $k$ time steps are performed in the GT mode, followed by decoding the rest in Prop mode. Note that we utilize this mode in both training as well as inference (test). This mode is a novel variant with benefits demonstrated in Section \ref{Sec:WellControlOptimizationResults}. 
\end{itemize}

\subsection{Experimental Evaluation}

\subsubsection{Datasets} \label{Sec:Datasets}
In order to develop various aspects of the neural network architecture, we have used the two BOS mentioned in Section \ref{Sec:ReservoirModelAndSimulation}, namely, OPM and Eclipse. 
While the OPM was used in majority of the tests, Eclipse was used to generate simulations with well control optimization - a feature not yet available in OPM.

Two main datasets were created: (a) 22k simulations via OPM (fixed well control), and (b) 32k simulations via Eclipse (well control optimized). Each such set was partitioned into training (TRAIN), validation (VALID), and test (TEST) sets, as shown in Table \ref{Tab:Datasets}, maintaining a proportion 80\%, 10\%, and 10\%, respectively. Additionally, a large dataset involving 163k OPM simulations (OPM-163k) was collected to examine effects of varying training set size on the resulting error metrics. 
\begin{table}[t]
  \centering
  \setlength\tabcolsep{2.2pt}
  \begin{tabular}{l|ccc}
  \toprule
  ~ & \multicolumn{3}{c}{Partition Size (Simulations)} \\
  Name & TRAIN & VALID & TEST \\
  \hline
  {\bf OPM-22k}        &17600 & 2200      & 2200 \\
  {\bf ECL-32k}		   & 28682  & 3585       & 3585  \\
  {\bf OPM-163k}        &130778 & 16347      & 16347 \\ 
  \hline
  \end{tabular}
  \caption{Datasets and their partitioning}
  \label{Tab:Datasets}
\end{table}
We want to emphasize that the OPM-163k only serves exploratory purposes of this paper and, in its size, would unlikely be a practical choice due to the considerable computational burden required to generate the corresponding simulations. 
In selected experiments (Section \ref{Sec:TrainingSizeResults}), the OPM-22k was further downsampled to investigate dependency on the training size.

Each simulation input consists of a uniform random choice over 500 realizations (see Section \ref{Sec:ReservoirModelAndSimulation}) and an action sequence of length 20. 
For each of the 20 actions, a decision to drill was made with 99\% probability, with a ratio $5:1$ in favor of drilling a producer. The well location follows a uniform random distribution with the constraint of not drilling within a 2-cell neighborhood of pre-existing wells. 

\subsubsection{Features and Preprocessing}
\subsubsection*{Actions and Realization} \label{Sec:ActionsRealizations}
The primary information to be encoded in the input $X = \{x_k\}_{0\leq k<K}$ are the drilling decisions (see Section \ref{Sec:Task}). Since the SPE9 reservoir has dimensions $24 \times 25 \times 15$ grid cells, initially, a joint action-location encoding was defined on a discrete set of 1201 (2x24x25+1) symbols. An alternative separate encoding for the action (3 symbols), the x-coordinate (24 symbols), and the y-coordinate (25 symbols), however, outperformed the joint encoding in our experiments, as will be shown below.

A realization ID associated with an input sequence is attached to each $x_k$ as a discrete variable. 

In order to provide the neural network with the above input, the discrete variables are "embedded" in a continuous space of certain dimensionality (a hyperparameter), similar to a method of embedding words in Natural Language Processing \cite{Mikolov_NIPS2013}. The embedding vectors for each discrete variable are initialized to random values and are trained jointly with the rest of the network. Negligible sensitivity was observed due to the dimensionality hyperparameter. Values we used are listed in Table \ref{Tab:FeatureDims}.
\begin{table}[t]
  \centering
  \setlength\tabcolsep{2.2pt}
  \begin{tabular}{l|c|c}
  \toprule
  Input & Type (Cardinality) & Dim \\
  \hline
  Well Type        & Discrete (3) & 3 \\
  Location X        & Discrete (24) & 10 \\
  Location Y        & Discrete (25) & 10 \\
  Realization        & Discrete (500) & 20 \\
  Geology        & Continuous & 20 \\
  \hline
  \end{tabular}
  \caption{Input features with their type and dimensionality}
  \label{Tab:FeatureDims}
\end{table}

\subsubsection*{Geological Features} \label{Sec:GeologicalFeatures}
Hypothesizing that local geological properties influence the flow through the wells the most, we added features related to the neighborhood of the well locations. 
More specifically, the following local (per-cell) features were considered: (a) rock type (shale, sandstone), (b) porosity, (c) permeability in horizontal and vertical direction. In our case, 5 cells at the bottom (for injectors) and 5 at the top (for producers) are affected by drilling, resulting in 20 real-valued features concatenated to form the final feature vector (see Table \ref{Tab:FeatureDims}).

\subsubsection*{Standardization}
Standardization was applied to the output variables before modeling, via a linear transform $y_{(k)}' = (y_{(k)} - \mu_k) / \sigma_k$ with $y_{(k)}$ being the $k$-th dimension of the output vector and $\mu_k, \sigma_k$ denoting the mean and standard deviation estimates obtained from the corresponding dimension over the TRAIN partition. The variables $y'$ are transformed back to their physical range using the inverse transform. All experimental evaluation is then performed on unnormed output. 

\subsubsection{Error Metric}
The core metric in our experimentation is the prediction error relative to target (GT) generated by the simulator. We base this metric on an L2 norm. Let $Y_k=\{y_{kt}\}_{0\leq t<T}$ and $\hat{Y}_{kt}=\{\hat{y}_{kt}\}_{0\leq t<T}$ denote target and predicted values for a simulation $k$ of length $T$. Let $K$ denote the total number of simulations in the test set. While the reservoir simulator produces rates at each time step, for practical use we are interested in the {\em cumulative} output. Hence, the cumulative value $z_{kt} = \sum_{\tau=0}^{t} y_{k\tau}$ is calculated forming a vector ${\bf z}_k = [z_{k0}, ..., z_{kT-1}]$ (and similarly for $\hat{z}_{kt}$).
The (relative) error of a simulation $k$ is then defined as follows
\begin{equation}
e_k = \frac{||\hat{\bf z}_k-{\bf z}_k||_2}{\overline{||{\bf z}||}_2}\label{Eq:RelErrorSim}
\end{equation}
Note that the denominator is calculated as an average over the entire test set. This mitigates issues with near-zero targets that occur in some valid cases. 
The final error rate used for reporting in this paper is the average of $e$ from Eq. \ref{Eq:RelErrorSim} over the entire test set.

\subsubsection{Training Procedure}
The system was implemented in Tensorflow \cite{TFpaper}. Each model was trained using the TRAIN partition (see Table \ref{Tab:Datasets}) to minimize the sequence MSE via the Adam optimizer \cite{AdamKingmaB14}, performed in batches of 100 simulations and an initial learning rate of 0.001. Parameters with the best loss on the VALID partition were then chosen. In case of GT Pre-Training (see Section \ref{Sec:DecodingVariants}), the resulting parameters served as starting point for the next training round with a slower learning rate of 0.0002. Trained models were evaluated using the TEST partition.

We experimented with varying model size in terms of the number of LSTM layers (ranging between 1 and 5), and hidden units (ranging between 32 and 2048). In the experimental sections below we report results on three representative configurations by memory footprint ("\#units x \#layers"): $\mbox{\texttt{1024x2}}$ (large), $\mbox{\texttt{128x5}}$ (medium), and $\mbox{\texttt{64x1}}$ (small).

\subsection{Baselines} \label{Sec:Baselines}

We consider several traditional techniques as references for comparison, e.g., upscaling \cite{Durlofsky2005}
offers itself as a suitable baseline to achieve acceleration due to reduced refinement and complexity (see below). We also describe two additional simple machine learning baselines as references.

\subsubsection{Upscaling} \label{Sec:Upscaling}
While oil reservoir fluids flow through microscopic pores, practical reservoir models, such as SPE9 in Figure \ref{Fig:SPE9Reservoir}, have homogeneous properties over tens of meters.
These coarser models are constructed attempting to approximate and analyze reservoir flow performance of multiscale nature with the available computational power.
To this end, upscaling consists of a set of procedures to obtain coarser reservoir models for flow performance prediction from geological characterizations which typically contain $10^7-10^8$ cells \cite{Durlofsky2005}.
Extrapolating from this idea, a reasonable question to ask is how the proxy models based on Neural Networks perform compared to even coarser versions of the base case reservoir model obtained with simple upscaling procedures.

We apply single-phase upscaling based on averaging reservoir properties aiming at a comparison of accuracy and performance with the proxies.
Given the base grid of $24 \times 25 \times 15$ grid-blocks we constructed two coarser grids of $12 \times 13 \times 15$ and $8 \times 9 \times 15$ grid-blocks, referred to as ``UP2'' and ``UP3,'' respectively.
The coarser grids average four $(2 \times 2 \times 1)$ and nine $(3 \times 3 \times 1)$ horizontal neighbor blocks in the ``UP2'' and ``UP3'' case, respectively.
All the relevant spatial reservoir properties, namely pressure, saturation, absolute permeability and porosity, are averaged on the pore-volume of the union of cells.
These averaging computations are performed in negligible computational time when compared to the solution of non-linear equations for simulation.
Finally, the coarser reservoir description is simulated with the BOS as usual.

While an increased error relative to the refined base model is anticipated, it should be pointed out that the upscaling procedure does not require any data to train or tune its parameters. In this respect, upscaling offers an advantage over learning proxies in use cases where there is an absolute lack of such training data. 

\subsubsection{Fixed Predictors}
Two simple baselines using fixed predictors are created as follows:
\begin{itemize}
\item A predictor that at all times outputs the mean value for each variable as seen in the training data (referred to as "Mean Baseline");
\item A predictor outputting the mean vector of all observations at the corresponding time step in the training data (referred to as "Time-Step Mean," or "TSM"). In other words, this predictor generates average flow curves for each component, as observed in the corresponding TRAIN partition. 
\end{itemize}
The two baselines above use the same training data as the proxy model and thus offer helpful calibration points. Obviously, the two fixed predictors above incur only negligible latency.

\subsection{Experimental Results}

\subsubsection{Results on OPM-22k} \label{Sec:ResultsOPM-22k}
Calculated according to Eq. \ref{Eq:RelErrorSim}, Table \ref{Tab:OPM-22k-MainResults} summarizes error rates for the essential techniques described in previous sections, using the $\mbox{\texttt{1024x2}}$ model.
For comparison, we also compute the simple baselines proposed in Section \ref{Sec:Baselines}.
``Mean Baseline'' and ``TSM Baseline'' have an error of 43.1\% and 39.3\%, respectively, whereas the upscaling cases ``UP2'' and ``UP3'' have an error of 19.1\% and 32.3\%, respectively.
Then, starting with the basic encoder-decoder at 21.5\%, pre-training a model in GT mode, followed by Prop training at a slower learning rate results in a significant decrease of error rate - by 6.2\% absolute to 15.3\%.
Replacing jointly encoded action-location input (see Section \ref{Sec:ActionsRealizations}) by factored action-location information and by adding geological features (see Section \ref{Sec:GeologicalFeatures}) decreases the error rate further to 12.2\%.
Finally, the attention mechanism attending to all encoder layers achieves an error rate of 10.3\%. For comparison we also give results for the more standard attention setup using the top layer which is inferior at 11.2\%. All differences are statistically significant with $p<0.001$ on a paired-sample t-test.
\begin{table}[t]
  \centering
  \setlength\tabcolsep{2.2pt}
  \begin{tabular}{l|c}
  \toprule
  Configuration & Sequence Error  \\
  \hline
  Mean Baseline & 43.1\% \\
  TSM Baseline & 39.3\% \\
  Upscaling UP3 & 32.3\% \\
  Upscaling UP2 & 19.1\% \\
    \hline
  Encoder-Decoder (1024x2)   & 21.5\% \\
  + GT Pre-Training    & 15.3\% \\
  + Factored Geol. Features    & 12.2\% \\
  (+ Single-Layer Attention) & 11.2\% \\
  + Multi-Layer Attention    & 10.3\% \\
  \hline
  \end{tabular}
  \caption{Sequence error rates of various configurations of the 1024x2 model on the OPM-22k TEST partition.}
  \label{Tab:OPM-22k-MainResults}
\end{table}
Later on, we refer to the multi-layer attentive, GT-pretrained models using a prefix $\mbox{\texttt{M}}$, e.g., as $\mbox{\texttt{M1024x2}}$. 

\begin{figure}[htbp!]
       \centering
       \includegraphics[height=3.5cm, width=0.44\textwidth]{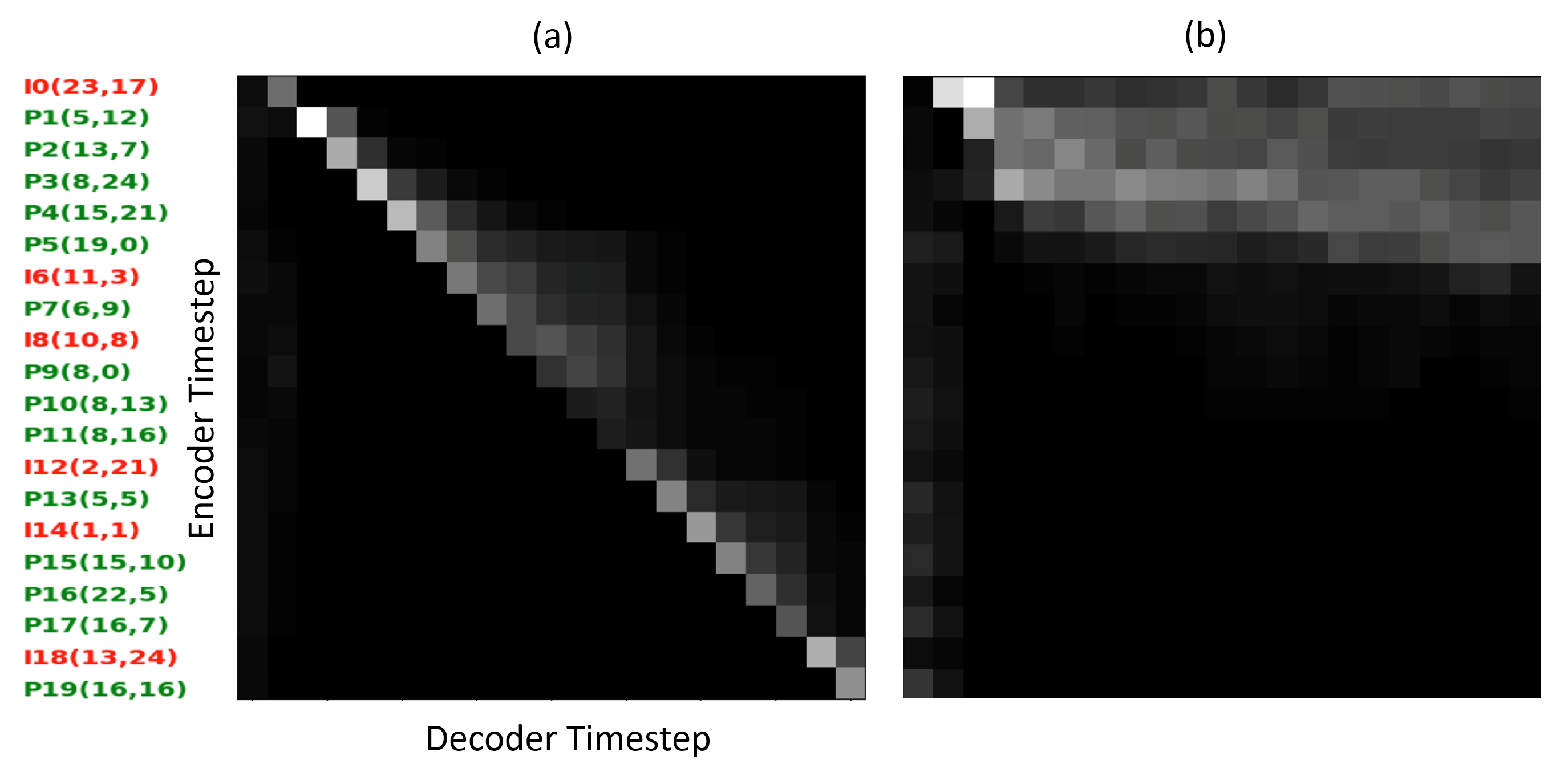}
       \caption{Attention masks visualized on first simulation from the TEST partition (sim=0): (a) Propagation decoding mode, (b) Ground Truth (GT) decoding mode}
       \label{Fig:AttentionMasks}
\end{figure}

In order to gain insight into how the attentive decoder focuses with respect to the input, Figure \ref{Fig:AttentionMasks} shows two example attention masks, i.e., the weighting distributions $\{\alpha_{kt}\}_{0\leq k<K}$ at each decoder time step $t$ over the input action space. In each panel, the horizontal axis corresponds to decoder time and the vertical axis to well drilling sequence (abbreviated as $\mbox{\texttt{\{I,P\}\#(x,y)}}$). It is interesting to note that while there is a fairly linear alignment between the decoder focus (light areas) and the input actions in the Prop mode (a), the bulk of attention is dispersed over an initial third of the input sequence when the decoder is given the GT at every time step (b). We hypothesize that the sharper action alignment emerges as the decoder obtains strong signal from the action sequence, while consuming its own, error-prone prediction as input. In contrast, the GT decoding mode supplies the decoder with accurate input making the decoder mostly rely on that, rendering the attention less important.

\begin{figure}[htbp!]
       \centering
       \includegraphics[height=5cm, width=1.0\linewidth]{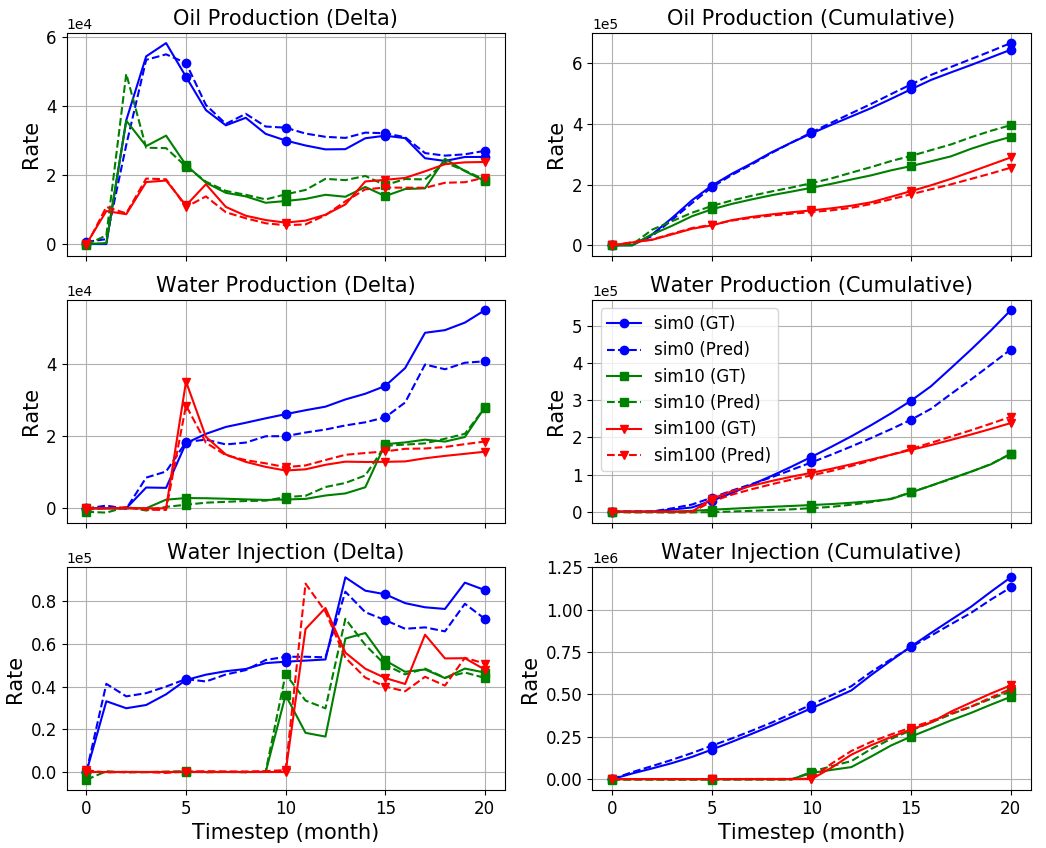}
       \caption{Production rates: Ground truth ("GT," solid), and prediction ("Pred," dashed) curves obtained from a M1024x2 model with attention on three randomly selected simulations. The model has an overall error rate of 10.3\%.}
       \label{Fig:ProductionRates}
\end{figure}

Figure \ref{Fig:ProductionRates} compares target (GT) and predicted production rates in three randomly selected simulations drawn from the TEST partition (sim: 0, 10, and 100). The first column in Figure \ref{Fig:ProductionRates} shows direct output of the model, i.e., rates (in barrels) for a given time period (30 days), while the second column compares the corresponding cumulative curves. The latter eventually serves the NPV calculation of the field as mentioned in Section \ref{Sec:Task}. 

\begin{figure}[htbp!]
       \centering
       \includegraphics[height=3.5cm, width=1.0\linewidth]{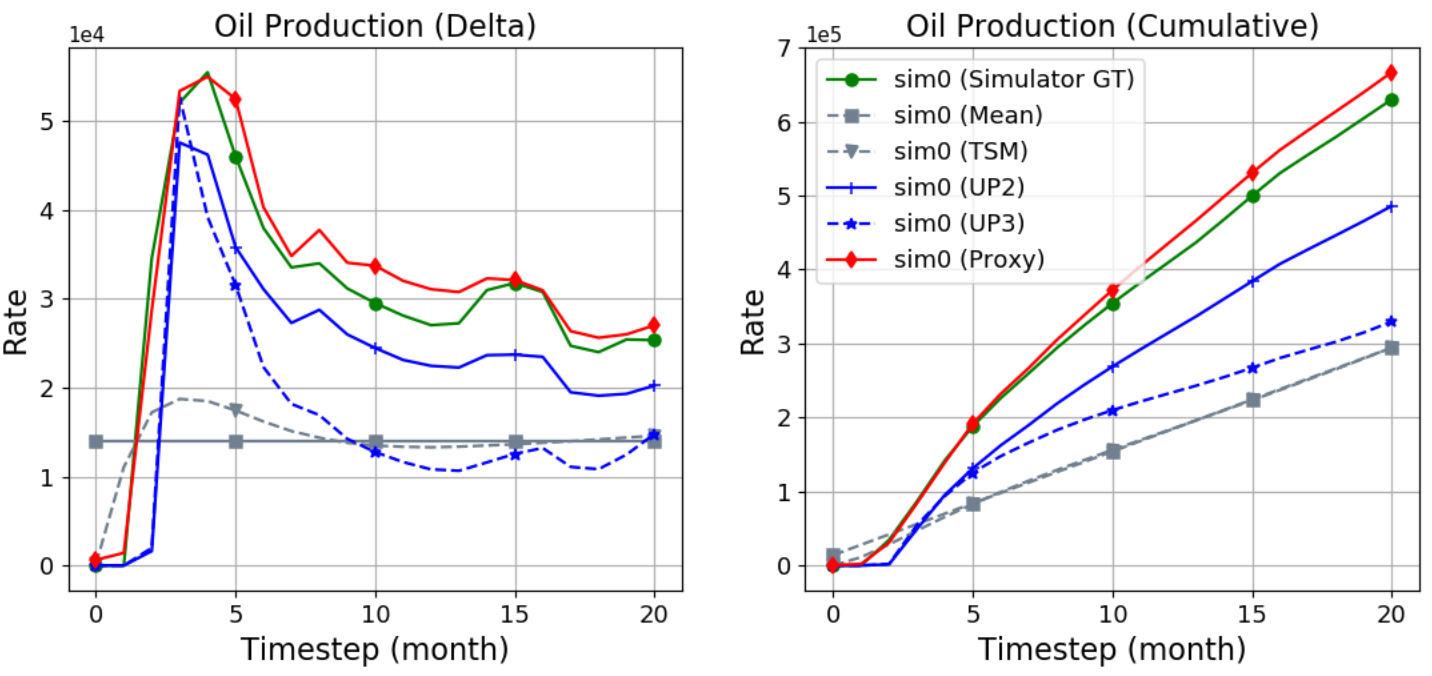}
       \caption{Predictions produced by the various baselines and the proxy in comparison with the simulator ground truth ("GT"). Shown are the field oil rates of a single simulation ("sim0"). }
       \label{Fig:ProductionRatesBaselines}
\end{figure}

Figure \ref{Fig:ProductionRatesBaselines} compares predicted oil production rates curves across the various baselines on an example of the first ("sim0") simulation, namely the "Mean," "Time-Slot-Mean (TSM)," upscaled "UP2," and "UP3" baselines as well as the proxy.
As can be seen, the proxy tends to approximate the ground-truth (GT) flow most accurately, followed by the UP2 baseline. The UP3, Mean, and TSM baselines tend to be considerably further off the target. 

\subsubsection{Varying Training Size} \label{Sec:TrainingSizeResults}
\begin{figure}[htbp!]
       \centering
       \includegraphics[height=5.5cm, width=1.0\linewidth]{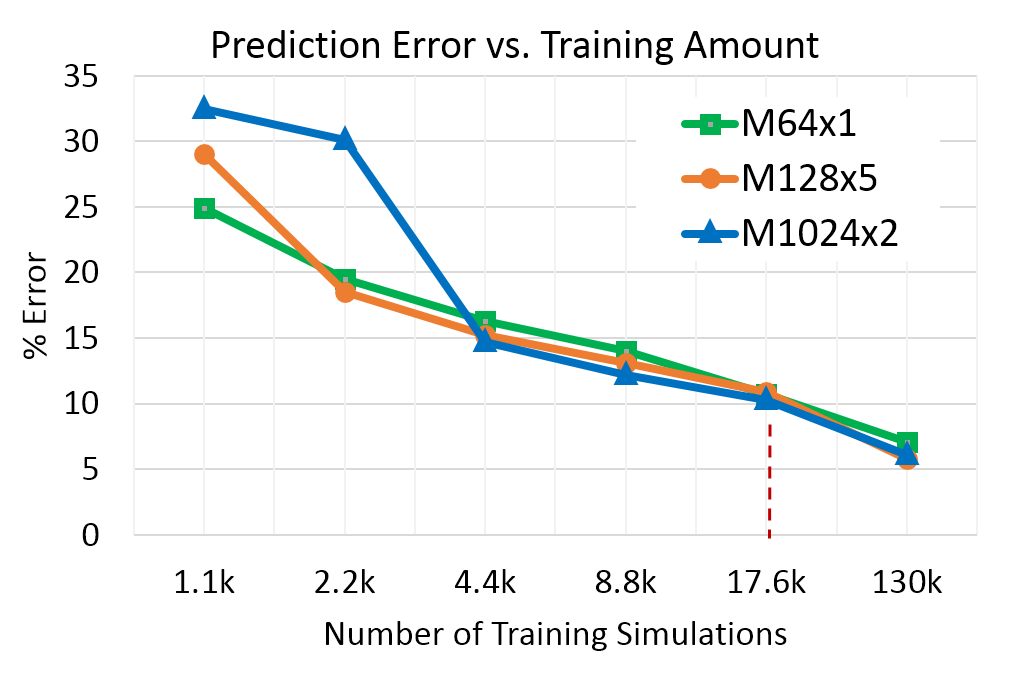}
       \caption{Prediction error as a function of training data amount. The dashed vertical line indicates training size of the OPM-22k dataset.}
       \label{Fig:OPMTrainingSizes}
\end{figure}

To assess the dependency of the prediction error on training data amount we conducted an experiment with subsampling the TRAIN partition to smaller amounts.  Figure \ref{Fig:OPMTrainingSizes} shows the results: starting from the main OPM-22k result (see above) with 17.6k simulations, the training amount was halved repeatedly to 1.1k simulations - an amount extremely small considering that there are 500 different reservoir realizations to be covered. On the other extreme, the training set was augmented by the remaining simulations from OPM-163k yielding a TRAIN partition with  about 130k simulations. In Figure \ref{Fig:OPMTrainingSizes}, two trends are evident: (1) models with higher parameter complexity tend to outperform lower-complexity ones for larger training amounts, with trend reversal at the lower end of the x-axis, and (2) there is a roughly linear relationship between the error rate and the log-scaled training amount. The accuracy of the rel. small $\mbox{\texttt{64x1}}$ model is remarkably competitive across the conditions. 

\subsubsection{Timing Comparison} \label{Sec:PerformanceComparison}
The main goal of the proxy is achieving significant acceleration compared to the physics-based simulation. Having established accuracy figures in the experiments above, we now turn to the question of performance (timing). We want to compare the average time needed to generate a simulation with 20 actions and 20 outputs, as above, between the OPM simulator and a selected set of proxy models. 
For the timing experiments we use OPM (2107.10 release) $\mbox{\texttt{ebos\_2p}}$ 
which we found to be the fastest implementation with an executable optimized for speed. Benchmarking was conducted on a server equipped with 4 x E7-4850 v2 CPUs @ 2.30 GHz (48/96 physical/virtual cores), 256 GB RAM, Open MPI implementation of MPI (for \texttt{mpirun}; version: 1.2.10) and standard \texttt{multiprocessing.Pool} objects (from Python, version: 2.7.14).
A comparison is somewhat complicated by the fact that (1) the OPM simulator is used as a black box and timing includes its initialization, and (2) the OPM code does not support the use of a GPU thus leaving only CPU comparison. (1) has a relatively minor effect as, judging from logging, the initialization stage of OPM takes a negligible fraction of the total time. 

Table \ref{Tab:TimingResults} summarizes the benchmarking results. We distinguish three devices of practical interest: (a) CPU 1-core, (b) CPU 8-core, and (c) single GPU (NVIDIA Tesla K80). All values in Table \ref{Tab:TimingResults} are durations (in milliseconds) of a simulation averaged over 100 measurements using different realizations. The first row shows the OPM time on a single core as well as an 8-core CPU (employing MPI), resulting in an average runtime of about 10 and 4.3 s/sim, respectively. The next three rows give the timing for the three model sizes benchmarked across the three devices. A clear advantage of the small-footprint $\mbox{\texttt{M64x1}}$ model emerges, in particular in the 1-Core case achieving a speedup factor of 2343X over the OPM simulator. The GPU seems to provide an advantage only in case of the largest model.

The last block of three rows shows simulation time in a case where 100 input sequences (simulation requests) can be batched up at once, thus allowing for the respective device to better utilize matrix operations. This sort of batching is of practical use in certain applications (e.g., Field Development Optimization using Monte Carlo methods). 
A significant speedup (0.1 ms/sim) can be seen now on the GPU side taking full advantage of its internal memory and architecture.

In a separate experiment, performance gains due to upscaling are assessed on a single core of an Intel Xeon CPU E5-2680 v2 @ 2.80GHz contrasted with Eclipse 2011.1 \cite{EclipseURL} (we experienced difficulties getting the desired upscaling setup working using OPM).
In this setting, the average running times of the OPM-22k TEST partition are  5125 (ms) when performing no upscaling, 1489 (ms) when performing the UP2, and 1155 (ms) the UP3 upscaling. 
Thus, the speed-up with upscaling ranges from 3X to 4X.

\begin{table}[htbp!]
  \centering
  \setlength\tabcolsep{2.2pt}
  \begin{tabular}{l|ccc}
  ~ & \multicolumn{2}{c}{CPU Device} & GPU Device \\
   Test & 1-Core & 8-Core & 1-GPU \\
   ~ & \multicolumn{3}{c}{Simulation Time (ms)} \\
  \hline
  OPM ebos\_2p        & 10310 &  4340    & n/a \\
  \hline
   M64x1 & 4.4  & 4.9  & 12.1  \\   
   M128x5 & 34.1   & 18.9  & 27.4  \\   
   M1024x2 & 299  &  330  & 31.6   \\
  \hline
   M64x1 Batch 100 & 2.9   & 0.6   & 0.1  \\   
   M128x5 Batch 100 & 32.2  & 5.1   & 0.6  \\   
   M1024x2 Batch 100 & 170 & 25.2 & 1.8  \\   
   \hline
  \end{tabular}
  \caption{Timing results. All values are averages over 100 measurements using same random set of simulations.}
  \label{Tab:TimingResults}
\end{table}

Another important (and advantageous) aspect of the proxy benchmarking results is their consistency. For instance, the $\mbox{\texttt{M64x1}}$ single-core result is $4.4 \pm 0.1$ ms/sim which is a range of $\pm 3\%$ relative to the mean. The corresponding OPM measurement of $10310 \pm 4030$ ms/sim exhibits a considerably larger range of $\pm 39\%$, which is typically caused by convergence issues for certain action sequences and realizations.
This large variability is also observed when running the upscaled cases.

Overall, considering the error patterns associated with each model size (see Figure \ref{Fig:OPMTrainingSizes}), the small proxy model $\mbox{\texttt{M64x1}}$ offers an acceptable error rate while achieving the highest speedups.

\subsubsection{Extrapolating to a Longer Simulation Horizon}
An interesting question arises regarding the proxy model's extrapolation capability. Suppose we are given a model trained with 20 input actions and 20 months worth of output. We now want to run the model for an extended period, say, 40 months. How well does such a model generalize in comparison to a model trained on 40 months worth of ground truth? To investigate, we generated simulations identical to those in OPM-22k but with output extended to 40 time steps, i.e. 1200 days, (labeled OPM-22k-E) allowing a direct comparison of such two models. Table \ref{Tab:ExtendedHorizonResults} summarizes the results in terms of average error rates, with $\mbox{\texttt{M128x5-E}}$ denoting a model trained on OPM-22k-E and $\mbox{\texttt{M128x5}}$ one trained on OPM-22k, as before. 
The first row in Table \ref{Tab:ExtendedHorizonResults} shows that both models perform comparably on OPM-22k, with $\mbox{\texttt{M128x5-E}}$ having no troubles to predict a shortened horizon of 20 time steps. The more interesting case of $\mbox{\texttt{M128x5}}$ extrapolating to 40 time steps is shown in the second row, where it achieves an error rate of 13.0\% - a moderately elevated error over the matched $\mbox{\texttt{M128x5-E}}$ at 10.6\%. 
While this increase is statistically significant, it seems to be sufficiently limited for us to conclude that the model has a reasonable capability to extrapolate to longer horizons not seen during its training. Furthermore, it is reassuring to observe the error rates of both models tested on the shorter time horizon perform equally well. This suggests that training on data with longer horizon is beneficial, whenever possible. 
\begin{table}[htbp!]
  \centering
  \setlength\tabcolsep{2.2pt}
  \begin{tabular}{l|c|c}
  \toprule
  Testset & M128x5 & M128x5-E  \\
  \hline
  OPM-22k     & 10.8\% & 11.1\% \\
  OPM-22k-E     & 13.0\% & 10.6\% \\
  \hline
  \end{tabular}
  \caption{Error rates of two models on the extended-time horizon (OPM-22k-E) and regular-horizon (OPM-22k) data with the corresponding models, $\mbox{\texttt{M128x5-E}}$ and $\mbox{\texttt{M128x5}}$.}
  \label{Tab:ExtendedHorizonResults}
\end{table}

\subsubsection{Modeling Individual Wells}
All experiments so far involved predicting {\em field} rates, i.e., total production and injection rates over all wells. In some applications, however, more detail may be needed. In one of our use cases (an optimization of a Field Development Plan) predictions of all producer wells were required. This corresponds to adding 20 new output variables each mapped to wells in the order of drilling (as there may be up to 20 producers). We then train the model using OPM-22k via same steps as before resulting in a model with 23-dimensional output. 

A comparison between the best wells model (using suffix "W", i.e., $\mbox{\texttt{M64x5+W}}$) and the best field-rates-only counterpart ($\mbox{\texttt{M1024x2}}$) is given in Table \ref{Tab:WellsModelResults}. Clearly, the task has become considerably more challenging as the wells model must maintain accuracy across 23 variables. Compared on the field rates only, the wells model is at 14\%, while the dedicated model is around 10\%, and at this error rate it still outperforms all other baselines. The error over the 20 well rates is at 19\% for the proxy compared to 16.6\% and 23.8\% for the UP2 and UP3 baselines, respectively.  
\begin{table}[t]
  \centering
  \setlength\tabcolsep{2.2pt}
  \begin{tabular}{l|c|c|c}
  \toprule
  ~ & \multicolumn{3}{c}{Error Rate (OPM-22k)} \\
  Model & Total & Field Rates & Well Rates  \\
  \hline
  TSM Baseline & 51.1\% & 39.3\% & 53.0\%\\
  UP3 Baseline & 24.9\% & 32.3\% & 23.8\%\\
  UP2 Baseline & 16.9\% & 19.1\% & 16.6\%\\
  \hline
  M64x5+W    & 18.3\% & 14.0\% & 19.0\%\\
  M1024x2          & 10.3\%   &  10.3\% & n/a  \\	  
  \hline
  \end{tabular}
  \caption{Error rates comparison between the wells model predicting 20 wells + 3 field rates and the best field-rates-only model on OPM-22k.}
  \label{Tab:WellsModelResults}
\end{table}
\begin{figure}[htbp!]
       \centering
       \includegraphics[height=2.5cm, width=0.8\linewidth]{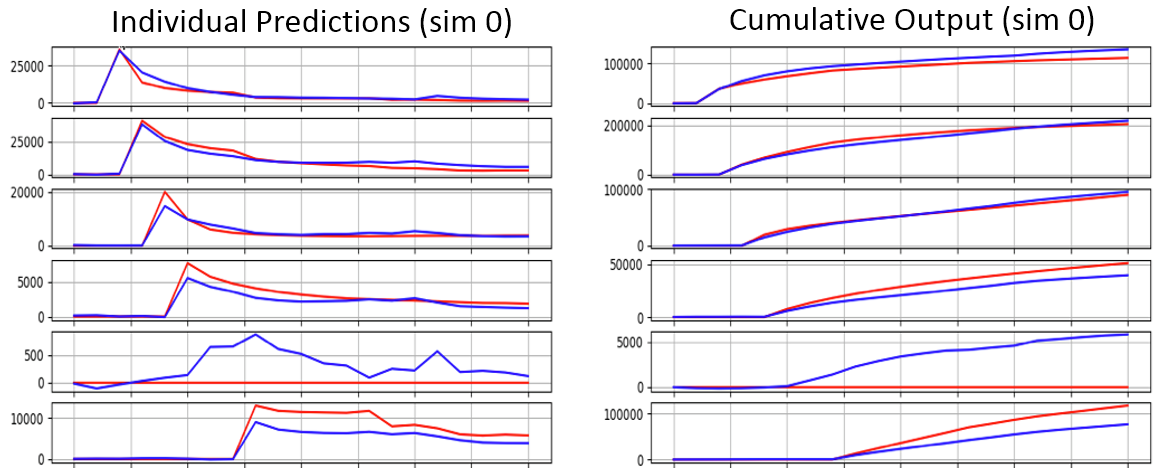}
       \caption{A sample of individual well (oil producer) predictions (blue) along with their ground truth (red).}
       \label{Fig:IndividualWellRates}
\end{figure}
Figure \ref{Fig:IndividualWellRates} shows a small sample of individual well predictions along with their ground truth. 
The output variables map to each producer in the order of drilling which corresponds to the timed onset of production visible in the cumulative curves. We observed that the model tends to overestimate wells that remain non-productive (zero output for entire simulation), albeit by relatively small amounts.

\subsubsection{Well Control Optimization} \label{Sec:WellControlOptimizationResults}
Optimizing the well controls is an enhancement currently only available in the commercial Eclipse simulator (see Section \ref{Sec:ReservoirModelAndSimulation}). 
The simulator runtime is typically increased by a factor of 5-10 due to the optimization step. %
We want to assess the ability of the proxy model to capture the optimization implicitly from control-optimized simulations. The dataset Ecl-32k (see Table \ref{Tab:Datasets}) serves this purpose. 

\begin{figure}[htbp!]
       \centering
       \includegraphics[height=4cm, width=0.9\linewidth]{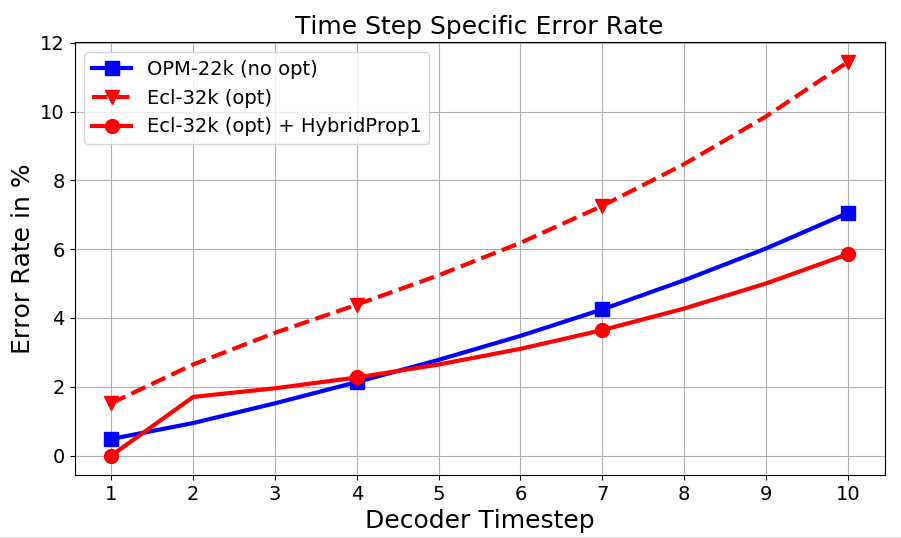}
       \caption{Effect of the HybridProp method on time step specific error rates, with the first time step being crucial.}
       \label{Fig:TimeSlotError}
\end{figure}

We start with a model $\mbox{\texttt{M128x5+W}}$ trained on Ecl-32k.
In comparison to levels seen with OPM-22k, the resulting error rates in the first row of Table \ref{Tab:HybridPropResults} lie considerably higher, at 27.1\%.
An inspection of the model output (see Figure \ref{Fig:TimeSlotError}, dashed curve) reveals a rel. high prediction error at time step 1 propagating further during decoding.
An in-depth variance analysis, which is omitted here due to space constraints, also reveals a high variability of the GT at time step 1 when (and only when) the optimization is present. This first step appears to be challenging for the decoder to predict accurately.
This observation motivated the idea behind HybridProp described in Section \ref{Sec:DecodingVariants}. In HybridProp, the decoder is given 1 or several GT frames as input both at training and test time to generate the rest of the sequence using regular propagation. 
An experimental assessment of this method is shown in Table \ref{Tab:HybridPropResults}. With $\mbox{\texttt{M128x5+W}}$ at 27.1\% error, giving the same model a single frame of GT during decoding only (referred to here as HybridDec) improves the error rate by 4.3\%. However, when the model is {\em retrained} with the same modification (HybridProp), the error declines dramatically, to 15.8\%. Providing more than one GT frame seems to yield diminishing returns. Despite the added requirement of having a simulator available at test time, this a remarkable and a practical result.
\begin{table}[t]
  \centering
  \setlength\tabcolsep{2.2pt}
  \begin{tabular}{l|c|c|c}
  \toprule
  ~ & \multicolumn{3}{c}{Error Rate (Ecl-32k)} \\
  Model & Total & Field Rates & Well Rates  \\
  \hline
  TSM Baseline & 51.0\% & 41.5\% & 52.4\% \\
  \hline
  M128x5+W        & 27.1\%   &  34.7\% & 29.9\%\\
  + HybridDec-1   & 22.8\%   &  27.5\% & 22.1\%  \\	  
  + HybridProp-1   & 15.8\%   &  21.3\% & 14.9\%  \\	  
  + HybridProp-2   & 15.4\%   &  21.0\% & 14.6\%  \\	  
  + HybridProp-4   & 15.1\%   &  21.5\% & 14.2\%  \\	  
  \hline
  \end{tabular}
  \caption{Error rates on Ecl-32k using a model without and with additional simulator input at prediction time.}
  \label{Tab:HybridPropResults}
\end{table}
Returning to Figure \ref{Fig:TimeSlotError}, after applying HybridProp1 the original (dashed red) curve moves down (solid red) to a trend comparable to one seen on non-optimized data (OPM-22k). 
The HybridProp method was also tested on models trained on OPM-22k and observed only negligible improvements further confirming our hypothesis regarding the high variance observed at first time step being unique to the well control optimization process.  

\section{Conclusion}\label{Sec:Discussion}
The series of experimental results presented in this paper demonstrates the effectiveness of the described proxy approach at accelerating reservoir model simulations, including variability due to uncertainty. We have observed a significant acceleration capability of more than 2000X compared to an industry-strength physics-based simulator OPM. Furthermore, we demonstrated it is possible to approximate the simulations with well control optimization thus offering an additional relative speedup. For practical amounts of training data, the accuracy of the neural network proxies presented here ranges between 10\% and 15\% error over 20-40 months horizons, relative to the simulator. The model shows a good extrapolation capability. Our proxy model captures non-linear interactions between wells, fluid, and rock, giving it a great advantage over state-of-the-art commercial techniques. When compared to simple upscaling procedures, we observe that the proxy models are capable to run about 500X faster and provide higher accuracy. We believe these outcomes, generated on a publicly available reservoir, are extremely promising and represent a valuable benchmark for future research in oil field development optimization. Moreover, we anticipate that, due to its application-agnostic nature, the approach is suitable for solving tasks in related fields of energy and environment modeling. 

\section {Acknowledgements}
The authors wish to thank Dr. Cristina Ib\'a\~nez of Repsol for her help generating well-control optimized simulations used in the Ecl-32k dataset. 

\bibliographystyle{ACM-Reference-Format}
\bibliography{our.bib}

\end{document}